\pdfoutput=1
\documentclass[10pt, a4paper]{article}
\usepackage{lrec-coling2024} 

\usepackage{adjustbox}
\usepackage{multirow}
\usepackage{enumerate}
\usepackage{expex}

\usepackage{CJKutf8} 

\newcommand{\Chinese}[1]{
  \begin{CJK*}{UTF8}{gbsn}
  #1 
  \end{CJK*} 
}
\newcommand{\Japanese}[1]{
  \begin{CJK*}{UTF8}{min}
  #1 
  \end{CJK*} 
}
\newcommand{\Korean}[1]{
  \begin{CJK*}{UTF8}{mj} 
  #1 
  \end{CJK*} 
}

\usepackage{tikz}
\newcommand*{\yellowhl}[1]{%
\tikz[baseline]\node[rectangle, fill=yellow, rounded corners, inner sep=0.3mm,anchor=base]{#1};%
}

\title{Grammatical Error Correction for Code-Switched Sentences \\by Learners of English}

\name{Kelvin Wey Han Chan\textsuperscript{1}, Christopher Bryant\textsuperscript{2,1}, Li Nguyen\textsuperscript{1}, \vspace{2mm} \\ \bf \large Andrew Caines\textsuperscript{1} and Zheng Yuan\textsuperscript{3,1} \vspace{2mm}} 

\address{
\textsuperscript{1} ALTA Institute \& Computer Laboratory, University of Cambridge, U.K. \\
\textsuperscript{2} Writer, Inc., San Francisco, U.S.A. \\
\textsuperscript{3} King's College London, U.K. \\
\texttt{firstname.lastname@cl.cam.ac.uk}\\
\texttt{kelvin@kelvinchanwh.com}\\
\texttt{zheng.yuan@kcl.ac.uk}
}

\abstract{
Code-switching (CSW) is a common phenomenon among multilingual speakers where multiple languages are used in a single discourse or utterance. Mixed language utterances may still contain grammatical errors however, yet most existing Grammar Error Correction (GEC) systems have been trained on monolingual data and not developed with CSW in mind. In this work, we conduct the first exploration into the use of GEC systems on CSW text. Through this exploration, we propose a novel method of generating synthetic CSW GEC datasets by translating different spans of text within existing GEC corpora. We then investigate different methods of selecting these spans based on CSW ratio, switch-point factor and linguistic constraints, and identify how they affect the performance of GEC systems on CSW text. Our best model achieves an average increase of 1.57 $F_{0.5}$ across 3 CSW test sets (English-Chinese, English-Korean and English-Japanese) without affecting the model's performance on a monolingual dataset. We furthermore discovered that models trained on one CSW language generalise relatively well to other typologically similar CSW languages.
 \\ \newline \Keywords{Code-switching, Grammatical error correction, Language learning} }

\hypersetup{
           breaklinks=true,   
           colorlinks=true,   
           pdfusetitle=true,  
        }

\begin{document}

\maketitleabstract

\section{Introduction}
\label{intro}

Code-switching (CSW) is a phenomenon where multilingual speakers use a combination of languages in a single discourse or utterance \citep{muysken_bilingual_2000,bullock_cambridge_2009}; this phenomenon is a common and natural practice in multilingual societies. Recent research has shown that CSW offers many pedagogical benefits, such as enhancing learners' communicative competence, linguistic awareness, and cultural identity \citep{Ahmad2009,Carsten2016,Wang2019,Daniel2019}. 

Grammatical Error Correction (GEC) meanwhile helps language learners by detecting and correcting various errors in text, such as spelling, punctuation, grammatical, and word choice errors. Most existing GEC systems, however, have been trained on monolingual data and not developed for CSW text. When given an input sentence, they therefore do not anticipate non-English words or phrases and so fail to detect and correct errors they otherwise normally resolve. This contrast is highlighted in examples (\ref{ex:cswError1}) and (\ref{ex:cswError2}), where a system was able to successfully resolve the missing verb error in the monolingual English sentence (\ref{ex:cswError1}), but was unable to resolve the same error in the equivalent CSW sentence (\ref{ex:cswError2}) where `pay' has been code-switched into Korean.

\pex 
\a \label{ex:cswError1}
\begingl[everygla=,everyglb=,aboveglbskip=-.2ex, aboveglftskip=.1ex]
\gla But the pay \textbf{\textcolor{red}{a}} little low .//
\glb \rightcomment{\textbf{[Monolingual English input]}} //
\glft `But the pay \textbf{\textcolor{Green}{is a}} little low .'//
\rightcomment{\textbf{[GEC System output]}}
\endgl

\a \label{ex:cswError2}
\begingl[everygla=,everyglb=,aboveglbskip=-.2ex, aboveglftskip=.1ex]
\gla But the \Korean{지불} \textbf{\textcolor{red}{a}} little low .//
\glb \rightcomment{\textbf{[Code-switching input]}} //
\glft `But the \Korean{지불} \textbf{\textcolor{red}{a}} little low .'//
\rightcomment{\textbf{[GEC System output]}}
\endgl
\xe

\noindent One major hurdle that directly affects the performance of GEC systems on CSW text is the lack of annotated CSW GEC datasets for training and testing \citep{nguyen_building_2022}. A common technique to compensate for the scarcity of large error-annotated corpora in GEC is to generate synthetic datasets (e.g.\ \citealp{yuan-etal-2019-neural,kiyono_empirical_2019,white_comparative_2020,stahlberg_synthetic_2021}). In this project, we similarly generate synthetic CSW GEC training data by translating different text spans within existing GEC corpora. We then evaluate our GEC systems trained using these synthetic datasets on three different natural CSW test datasets including English-Chinese (EN-ZH), English-Korean (EN-KO) and English-Japanese (EN-JA). We make our source code publicly available to aid reproducibility.\footnote{\url{https://github.com/kelvinchanwh/csw-gector}
}

This paper makes the following contributions:
\begin{enumerate}
    \item We conduct the first investigation into developing GEC models for CSW input. 
    \item We propose a novel method of generating synthetic CSW GEC data using a standard GEC dataset and a translation model.
    \item We introduce three new CSW GEC datasets to evaluate our proposed models. 
    \item We explore different methods of selecting text spans (with varying levels of involvement in linguistic theories) for synthetic CSW generation, and evaluate how this affects GEC performance for EN-ZH, EN-KO and EN-JA CSW text.
    \item We investigate the cross-lingual transferability of our models to CSW languages that they have not been trained on.
\end{enumerate}

\noindent Finally, it is worth making clear that we use the term `code-switching' in this work to encompass all instances of language mixing between English and non-English. We are aware of the longstanding debate between code-switching and borrowing in Linguistics \citep{muysken_bilingual_2000,nguyen_borrowing_2018,poplack_borrowing_2018,Deuchar2020,Treffers2023}, but this falls outside the scope of our focus and we hence do not make a distinction.

\section{Related Work}
\label{related}

\subsection{Synthetic GEC Dataset Generation}
\label{related:gec:synthetic}
Synthetic GEC dataset generation is a well-established practice \citep[Sec.~5]{bryant_grammatical_2023}. Various methodologies for generating synthetic data include the use of noise injection \citep{rozovskaya_generating_2010, felice_generating_2014, xu_erroneous_2019}, back translation \citep{rei_artificial_2017, yuan-etal-2019-neural, stahlberg_seq2edits_2020}, and round-trip translation \citep{madnani_exploring_2012, lichtarge_corpora_2019}. These techniques were all developed for monolingual GEC, however, and so are not directly applicable to CSW GEC.

\subsection{Synthetic CSW Generation}
\label{related:cs:gen}
The two main approaches to generating synthetic CSW text are linguistic-driven approaches and machine translation.

\paragraph{Linguistically Driven Approaches}
In linguistically-driven approaches, sections of text are commonly replaced based on intuitions derived from linguistic theories. For example, the Equivalence Constraint \citep{poplack_syntactic_1978} proposes that well-formed code-switching requires grammatical constraints to be satisfied in both languages. With this in mind, \citet{pratapa_language_2018} and \citet{pratapa_comparing_2021} generate Hindi-English CSW data by using the parse trees of parallel sentences to match the surface order of the child nodes.

In contrast, the Matrix Language Framework (MLF) \citep{myers-scotton_contact_2002} proposes an asymmetrical relationship between the languages in CSW sentences. Specifically, the MLF hypothesises that the matrix (i.e.\ dominant) language provides the frame of the sentence by dictating a certain subset of the grammatical morphemes and word order, while the embedded language only provides syntactic elements with little to no grammatical function \citep{johanson_dynamics_1999, myers-scotton_multiple_2005}. \citet{lee_linguistically_2019}, \citet{gupta_semi-supervised_2020} and \citet{rizvi_gcm_2021} thus developed tools to generate CSW text based on this principle. 

Finally, the Functional Head Constraint \citep{belazi_code_1994} posits that the strong relationship between a functional head and its complement makes it impossible to switch languages between these two constituents. \citet{li_code-switch_2012} used this constraint to generate CSW text by using a translation model to expand the search network and then parsing each possibility to filter for sentences permissible under the constraint. 

\paragraph{Machine Translation}
Machine translation was only recently tested on code-switching data \citep{nguyen-etal-2023-effective,nguyen-IJM-2023}, but was first used to generate CSW text by \citet{xu_can_2021}. In these systems, models are trained to treat English as the source language and CSW text as the target language. Recently, there has been an uptake in the use of Machine Translation as a method to generate CSW text due to the introduction of a large-scale code-mixed English-Hindi parallel corpus \citep{srivastava_hinge_2021} and the introduction of shared tasks for generating English-Hindi, English-Spanish and English-Arabic CSW text \citep{srivastava_overview_2022, chen_calcs_2022}.

\section{CSW GEC Dataset Generation}
\label{proposed:data}

\begin{table*}[!t]
  \centering
    \begin{tabular}{ll}
    \hline
    \textbf{Step} & \textbf{Sentence} \\
    \hline
    1. Input monolingual GEC data & What if \textbf{\textcolor{red}{human}} use up all the \textbf{\textcolor{red}{resource}} in the world? \\
          & What if \textbf{\textcolor{Green}{humans}} use up all the \textbf{\textcolor{Green}{resources}} in the world? \\
    \hline
    2. Select span & What if \textbf{\textcolor{Green}{humans}} use up all the \textbf{\textcolor{Green}{resources}} in the \yellowhl{world}? \\
    3. Translate span & What if \textbf{\textcolor{Green}{humans}} use up all the \textbf{\textcolor{Green}{resources}} in the \yellowhl{\Chinese{世界}}? \\
    4. Apply errors & What if \textbf{\textcolor{red}{human}} use up all the \textbf{\textcolor{red}{resource}} in the \yellowhl{\Chinese{世界}}? \\
    \hline
    5. Output CSW GEC data & What if \textbf{\textcolor{red}{human}} use up all the \textbf{\textcolor{red}{resource}} in the {\Chinese{世界}}? \\
          & What if \textbf{\textcolor{Green}{humans}} use up all the \textbf{\textcolor{Green}{resources}} in the {\Chinese{世界}}? \\
    \hline
    \end{tabular}
  \caption{Example CSW GEC data generation pipeline.}
  \label{tab:generation}
\end{table*}

\subsection{Data Augmentation Method}
The main idea behind our CSW GEC data generation method is that we select spans of tokens from the corrected side of existing GEC corpora and replace them with their tokenised translated equivalents. Specifically, we use the Google Translate API via the \verb|py-googletrans| package\footnote{\url{https://github.com/ssut/py-googletrans}} for all translations, and tokenise the Chinese, Korean and Japanese output respectively with Jieba,\footnote{\url{https://github.com/fxsjy/jieba}} Nagisa,\footnote{\url{https://github.com/taishi-i/nagisa}} and Komoran.\footnote{\url{https://github.com/shineware/PyKOMORAN}} Once the corrected portion of the GEC dataset has been converted to CSW text, the errors from the original sentences are then reapplied to the CSW corrected sentences.
This results in a dataset of tokenised CSW sentences which preserve the original human-annotated GEC errors in the source corpus. An overview of this process is shown in \autoref{tab:generation}.

\begin{table*}[t!]
  \centering
    \begin{tabular}{ll}
    \hline
    \textbf{Source} & She was going to have so many answers to so many questions .  \\
    \hline
    \textbf{ratio-token} &  She was \yellowhl{\Japanese{行きます}} to have so many \yellowhl{\Japanese{答え に}} so many questions . \\
    \textbf{cont-token} & She was going to have so many \yellowhl{\Japanese{そうへの答え}} many questions .  \\
    \textbf{rand-phrase} & She was going to have so many answers to \yellowhl{\Japanese{非常に多くの質問}} . \\
    \textbf{ratio-phrase} & She was going to have \yellowhl{\Japanese{非常に多くの答え}} to so many questions . \\
    \textbf{overlap-phrase} & She \yellowhl{\Japanese{非常に多くの質問に非常に多くの答えがあるでしょう}} . \\
    \textbf{noun-token} & She was going to have so many \yellowhl{\Japanese{答え}} to so many questions . \\
    \hline
    \end{tabular}
  \caption{Example output of different generation methods for English-Japanese (EN-JA)} 
  \label{tb:genexamples}
\end{table*}

\subsection{Span Selection Methods}
\label{methodology:data}
There are many ways to select different spans of text when generating CSW data. In this paper, we report six different span selection methods, namely \textbf{ratio-token}, \textbf{cont-token}, \textbf{rand-phrase}, \textbf{ratio-phrase}, \textbf{overlap-phrase}, and \textbf{noun-token}. Since one of our main objectives is to compare and contrast different methods of generating code-switched text for GEC (cf. \textsection \ref{intro}), we consider both naive options (\textbf{ratio-token}, \textbf{cont-token}) and linguistically motivated options (\textbf{rand-phrase}, \textbf{ratio-phrase}, \textbf{overlap-phrase}, \textbf{noun-token}). These variations were crucial for understanding the nuances of code-switching in the context of GEC.
We describe each of these methods in detail below. 

\paragraph{Ratio of code-switched tokens (ratio-token)}
\label{methodology:data:fractoken}
In the \textbf{ratio-token} method, we randomly sampled and translated tokens from the English source sentence until approximately 20\% of all tokens in the sentence were non-English. This ratio is not linguistically motivated, but set based on a qualitative analysis of CSW sentences in the multilingual Lang-8 learner corpus \citep{mizumoto_mining_2011}.

\paragraph{Ratio of continuous code-switched tokens (cont-token)} 
\label{methodology:data:contfractoken}
The \textbf{cont-token} method is based on the observation in the Lang-8 CSW dataset that speakers tend to code-switch from one language to another and back only once within a single sentence. Therefore, instead of selecting random tokens until we hit a target CSW ratio (i.e. $\sim$20\% as in \textbf{ratio-token}), we randomly select a starting point within the sentence and translate the following $n$ tokens until the CSW ratio is satisfied. Note that this differs from what we see in a speech-based context where shifts of topics and interlocutors might trigger more inter-sentential switches (\citealp{nguyen2021thesis,Gardner-Chloros_2009,muysken_bilingual_2000}, i.a.).

\paragraph{Random Phrase (rand-phrase)}
\label{methodology:data:randomphrase}
Linguistic research has also shown that CSW is usually based on a complete syntactic unit \citep{poplack_syntactic_1978,myers-scotton_contact_2002}. The \textbf{rand-phrase} method thus uses the Berkeley Neural Parser (\verb|benepar|) \citep{kitaev_constituency_2018, kitaev_multilingual_2019} to first identify syntactic phrases within the sentence, and then randomly selects one to be translated. Unlike the previous two methods, this ensures the CSW fragment is more linguistically plausible. 

\paragraph{Ratio of code-switched phrases (ratio-phrase)} 
\label{methodology:data:fracphrase}
The \textbf{ratio-phrase} method is similar to the \textbf{rand-phrase} method in that it relies on \verb|benepar| to first identify the phrases. Where it differs is that it aims to select and translate a phrase that has a length closest to the number of tokens required to meet the target CSW ratio (rather than select a phrase randomly).

\paragraph{Least overlap with edit spans (overlap-phrase)}
\label{methodology:data:intersectphrase}
All the above methods discard any errors that overlap with the randomly selected spans. This results in fewer training examples, and so the \textbf{overlap-phrase} method is designed to preserve as many edits as possible by selecting the longest phrase that minimally intersects with any edits.  
\paragraph{Code-switched noun tokens (noun-token)} 
\label{methodology:data:nountoken}
Finally, one of the few universal findings in linguistic research on code-switching is that a majority of natural CSW only involves a single noun token (\citealp{myers-scotton_duelling_1997, muysken_bilingual_2000, myslin_code-switching_2015,nguyen_borrowing_2018}, i.a.). The \textbf{noun-token} method thus leverages this insight by randomly selecting a single token with a \verb|NOUN| or \verb|PROPN| POS tag.\footnote{POS tagged using spaCy: \url{https://spacy.io}}

\autoref{tb:genexamples} provides a worked example of how different methods generate different synthetic CSW outputs on the same source sentence.

\section{Experimental Setup}
\label{proposed}

We use GECToR \citep{omelianchuk_gector_2020} as our baseline model.\footnote{\url{https://github.com/grammarly/gector}} GECToR is a sequence labeling approach that assigns an edit tag to each input word, where each edit tag represents the transformation that needs to be applied to correct the error; e.g. \texttt{\$KEEP} (no error), \texttt{\$APPEND\_the} (insert `the'), \texttt{\$NOUN\_NUMBER\_PLURAL} (make noun plural).\footnote{See \citet{omelianchuk_gector_2020} for the full list of tags.} The GECToR training process consists of three stages:

\begin{enumerate}
    \item Stage 1 (9m sentences) is trained on the synthetic sentences in the PIE dataset \citep{awasthi_parallel_2019}.
    \item Stage 2 (619k sentences) is trained on only the sentences that contain errors in the concatenation of NUCLE \citep{dahlmeier_building_2013}, the FCE \citep{yannakoudakis_new_2011}, the Lang-8 Corpus of Learner English \citep{tajiri_tense_2012}, and W\&I + LOCNESS \citep{bryant_bea-2019_2019}. 
    \item Stage 3 (34k sentences) is again trained on W\&I + LOCNESS \citep{bryant_bea-2019_2019} but this time includes sentences that do not contain any errors (i.e., the full training set). 
\end{enumerate}

\noindent We modify this setup in the following ways. First, we pass the data from each stage through our synthetic CSW data generation pipeline (using each span selection method) to introduce CSW fragments in all the input training sentences. This allowed us to investigate which synthetic data generation method yielded the most improvement. Since our preliminary experiments determined that the full three-stage training process would take about 18 hours on an Nvidia A100 GPU, our first experiment focused only on using our synthetic CSW data in stages 2 and 3 in order to reduce the amount of required computation. We later extended this setup to stage 1 once we knew the most promising configurations.

Second, we use XLM-RoBERTa \citep{conneau_unsupervised_2020} as our pretrained base model, as initial experiments showed it yielded the largest improvements compared to other pretrained models. This finding is consistent with \citet{winata_are_2021}, who similarly found XLM-RoBERTa performed best among multilingual models when predicting POS tags and named entities in CSW texts. We nevertheless note that other multilingual models such as mBERT,\footnote{\url{https://github.com/google-research/bert/blob/master/multilingual.md}} yielded similar improvements, while monolingual models such as BERT \citep{devlin_bert_2019}, RoBERTa \citep{liu_roberta_2019}, DeBERTa \citep{he_deberta_2021} and ELECTRA \citep{clark_electra_2020} were consistently worse. We suspect this is because for the languages we worked with, monolingual models treat all CSW tokens as out-of-vocabulary tokens. 

In all experiments, we use the default GECToR values of 0 for the \verb|additional_confidence| and \verb|minimum_error_probability| precision/recall trade-off inference parameters.

\section{Evaluation}
\label{evaluation}

We evaluate all our models using the standard ERRor ANnotation Toolkit (ERRANT) $F_{0.5}$ metric \citep{bryant_automatic_2017}. ERRANT automatically aligns parallel original/corrected sentence pairs and extracts and classifies the edits using a linguistically-enhanced rule-based approach. The extracted system hypothesis edits are then compared against the reference edits to calculate precision, recall and $F_{0.5}$. Since there are no previous CSW GEC test sets, we introduce our own benchmark, the Lang-8 CSW test set, which is based on the Lang-8 Learner corpus.

\subsection{Lang-8 CSW Test Set}
The Lang-8 Learner corpus is a large multilingual corpus containing forum posts and responses written by international language learners seeking help from native speakers online \citep{mizumoto_mining_2011}. A small fraction of these posts also contain natural code-switching sentences, which we identified using Google's Compact Language Detector.\footnote{\url{https://github.com/google/cld3}} Having extracted the sentences that contained English and exactly one other language, we applied some simple filters to reduce noise. Specifically, we removed sentences:

\begin{enumerate}[i.]
    \item that had no corrections;
    \item where the original sentence exactly matched the start of the corrected sentence; and
    \item where the length difference between the original and corrected sentence was more than 5 tokens.
\end{enumerate}

\noindent The first and second filters ensured we did not include sentences that were either unannotated or already correct,\footnote{As Lang-8 was not professionally annotated, users commonly add phrases like ``Well done!'' and ``Keep practising!'' as comments.} while the third filter ensured we did not include sentences that contained too many additional comments or other irrelevant strings. This resulted in a dataset of 201 English-Chinese (EN-ZH) sentences, 764 English-Korean (EN-KO) sentences, and 4,808 English-Japanese (EN-JA) sentences. CSW sentences from other languages were very rare and therefore excluded.

\subsection{Human Re-annotated Dataset}
Since the GEC annotations in the Lang-8 dataset were not created by professional annotators, they have a tendency to be noisy or incomplete. To mitigate this, we asked two bilingual annotators to reannotate a random selection of 200 English-Chinese and 200 English-Korean sentences respectively.\footnote{We were unable to recruit a bilingual English-Japanese annotator for the English-Japanese data.} Specifically, the English-Chinese sentences were annotated by a native Chinese speaker with English as a second language, and the English-Korean sentences were annotated by a native English speaker with Korean as a second language. In accordance with other professionally annotated datasets, the annotators were instructed to make minimal edits to the text \citep{bryant_grammatical_2023}, and also flag sentences that they were unable to correct or were sentence fragments; these fragments were later excluded from the human-annotated dataset (8 in EN-ZH and 16 in EN-KO). Additionally, annotators were asked to only correct the English tokens; if the non-English tokens contained grammatical errors, they were left uncorrected. 

\subsection{Test Set Distributions}
\label{datasets:testset}
The above steps resulted in the Lang-8 CSW test set, which contains three language pairs (EN-ZH/KO/JA) and a subset of high-quality human annotations. Various statistics about this dataset are shown in \autoref{tab:cs_ratio}.

\begin{table}[!t]
    \centering
    \begin{tabular}{llrrrrr}
    \cline{4-7}
        \multicolumn{2}{c}{} &  & \multicolumn{2}{c}{CSW (\%)} & \multicolumn{2}{c}{SPF}  \\
        \hline
        \multicolumn{2}{c}{Test set} & Sents & $\mu$ & $\sigma$ & $\mu$ & $\sigma$  \\ \hline
        ~ & ZH & 201 & 6.64 & 5.68 & 1.88 & 1.61 \\ 
        L8 & KO & 764 & 13.13 & 11.42 & 2.45 & 1.66 \\ 
        ~ & JA & 4,808 & 10.13 & 8.26 & 2.94 & 2.24 \\ \hline
        \multirow{2}{*}{HR} & ZH & 192 & 8.09 & 4.84 & 2.42 & 1.37 \\ 
        ~ & KO & 184 & 16.73 & 10.89 & 2.44 & 1.54 \\ \hline
    \end{tabular}
    \caption{Number of sentences, ratio of CSW tokens (mean and standard deviation) and switchpoint factor (SPF) (mean and standard deviation) for the Lang-8 (L8) and Human Re-annotated (HR) Test sets.} 
    \label{tab:cs_ratio}
\end{table}

\begin{table*}[!t]
    \centering
    \begin{tabular}{lllccccc}
    \hline
     \multicolumn{3}{c}{Training Dataset} & \multicolumn{3}{c}{Lang-8 CSW test set} & \multicolumn{2}{c}{Re-annotated CSW test set} \\ \hline
     Stg. 2 & Stg. 3 & Method & ZH & KO & JA & ZH & KO \\ \hline
    EN & EN & baseline & $32.95_{0.42}$ & $33.11_{0.03}$ & $28.60_{0.24}$ & $43.70_{1.06}$ & $23.82_{0.14}$ \\ \hline
    \multirow{6}{*}{EN} & \multirow{6}{*}{CSW} & ratio-token & $\mathbf{34.19_{0.73}}$ & $32.51_{1.02}$ & $\mathbf{29.28_{0.31}}$ & $\mathbf{43.76_{0.57}}$ & $\mathbf{24.95_{0.68}}$ \\ 
     &  & cont-token & $\mathbf{33.82_{0.44}}$ & $32.06_{1.09}$ & $\mathbf{28.86_{0.15}}$ & $43.30_{1.29}$ & $23.40_{0.73}$ \\
     &  & rand-phrase & $\mathbf{33.53_{0.53}}$ & $\mathbf{34.05_{0.57}}$ & $28.16_{0.69}$ & $\mathbf{44.93_{0.84}}$ & $\mathbf{24.58_{0.39}}$ \\
     &  & ratio-phrase & $\mathbf{34.17_{0.46}}$ & $32.40_{0.02}$ & $27.90_{0.51}$ & $\mathbf{43.84_{1.07}}$ & $23.17_{0.81}$ \\
     &  & overlap-phrase & $32.91_{0.72}$ & $31.35_{0.51}$ & $27.37_{0.63}$ & $40.87_{0.45}$ & $23.40_{0.40}$ \\
     &  & noun-token & $\mathbf{33.69_{0.34}}$ & $\mathbf{33.32_{1.02}}$ & $\mathbf{28.90_{0.21}}$ & $\mathbf{44.70_{0.46}}$ & $\mathbf{24.82_{1.20}}$ \\ \hline
    \multirow{6}{*}{CSW} & \multirow{6}{*}{EN} & ratio-token & $25.25_{3.04}$ & $30.51_{5.43}$ & $\mathbf{28.86_{0.43}}$ & $37.21_{1.72}$ & $22.44_{5.12}$ \\
     &  & cont-token & $29.22_{7.24}$ & $29.72_{3.95}$ & $28.27_{1.79}$ & $41.27_{7.46}$ & $21.08_{3.24}$ \\
     &  & rand-phrase & $\mathbf{33.72_{1.24}}$ & $\mathbf{34.18_{1.52}}$ & $\mathbf{28.67_{0.99}}$ & $\mathbf{45.59_{1.03}}$ & $\mathbf{25.87_{1.14}}$ \\
     &  & ratio-phrase & $\mathbf{34.07_{0.51}}$ & $31.75_{0.49}$ & $27.98_{0.63}$ & $\mathbf{45.98_{0.85}}$ & $23.58_{0.21}$ \\
     &  & overlap-phrase & $\mathbf{34.34_{0.35}}$ & $32.23_{0.32}$ & $28.53_{0.35}$ & $\mathbf{45.12_{0.55}}$ & $23.72_{0.54}$ \\
     &  & noun-token & $\mathbf{33.88_{1.13}}$ & $32.88_{0.15}$ & $\mathbf{29.10_{0.33}}$ & $\mathbf{46.09_{0.94}}$ & $\mathbf{25.32_{0.36}}$ \\ \hline
    \multirow{6}{*}{CSW} & \multirow{6}{*}{CSW} & ratio-token & $32.16_{1.84}$ & $30.10_{6.48}$ & $\mathbf{28.76_{0.39}}$ & $41.41_{4.41}$ & $\mathbf{23.92_{6.78}}$ \\
     &  & cont-token & $\mathbf{34.82_{0.39}}$ & $28.82_{4.87}$ & $27.53_{1.37}$ & $\mathbf{45.28_{1.12}}$ & $21.06_{5.02}$ \\
     &  & rand-phrase & $\mathbf{33.99_{2.08}}$ & $\mathbf{34.42_{1.61}}$ & $28.22_{0.82}$ & $\mathbf{46.28_{1.70}}$ & $\mathbf{25.35_{0.39}}$ \\
     &  & ratio-phrase & $\mathbf{34.40_{1.50}}$ & $30.76_{0.15}$ & $27.16_{0.55}$ & $\mathbf{45.27_{0.86}}$ & $22.76_{0.60}$ \\
     &  & overlap-phrase & $\mathbf{33.95_{1.01}}$ & $32.15_{0.68}$ & $28.23_{0.39}$ & $41.13_{1.24}$ & $22.06_{0.55}$ \\
     &  & noun-token & $\mathbf{33.67_{0.20}}$ & $\mathbf{33.24_{0.45}}$ & $\mathbf{29.04_{0.32}}$ & $\mathbf{46.23_{0.38}}$ & $\mathbf{24.88_{0.75}}$ \\ \hline

    \end{tabular}
    \caption{Table showing the $F_{0.5}$ score and the standard deviation for the different span selection methods using the XLM-RoBERTa model. Each test set was evaluated using models trained on CSW datasets in their respective languages. The $F_{0.5}$ score was averaged across three seeds. The language codes represent the different portions of the test dataset containing CSW text. Scores in bold indicate instances where the span selection method resulted in an $F_{0.5}$ score greater than the baseline.}
    \label{tab:data_aug_xlm}
\end{table*}

Specifically, the CSW ratio is the average ratio of non-English tokens per utterance, while the switchpoint factor (SPF) is the average number of times a speaker switches from one language to another in a sentence. For example, a CSW ratio of 6.64\% for Lang-8 CSW ZH indicates that we expect 6.64\% of all tokens in a sentence to be Chinese and the remaining 93.36\% to be English. We can see that the CSW ratio varies significantly between test sets, and in fact the Korean (KO) dataset has twice the number of CSW tokens than the Chinese (ZH) dataset. In contrast, datasets have an average SPF of $\sim$2.4, which indicates most sentences switch languages only once or twice. Incidentally, an odd numbered SPF indicates a sentence starts or ends with a non-English component, but most switches go from English to non-English then back again.

\section{Results \& Discussion}
We evaluate our models primarily on the aforementioned test sets and carry out three main experiments to: 
\begin{enumerate}[i.]
    \item compare different span selection methods (\textsection \ref{results:data});
    \item extend the best augmentation method to different training stages, including stage 1 (\textsection \ref{results:stage1}); 
    \item examine the effect of cross-lingual transfer; i.e. training on one CSW language and testing on another (\textsection \ref{results:language_inter}). 
\end{enumerate}

\subsection{Span Selection Method}
\label{results:data}
Results comparing the effect of each span selection method are presented in \autoref{tab:data_aug_xlm}. 

\begin{table*}[!t]
    \centering
    \begin{tabular}{lllllccc}
    \hline
        \multicolumn{2}{c}{Test} & \multicolumn{3}{c}{Training Dataset}  & \multirow{2}{*}{Prec} & \multirow{2}{*}{Rec} & \multirow{2}{*}{$F_{0.5}$}  \\ 
        \multicolumn{2}{c}{Dataset} & Stage 1 & Stage 2 & Stage 3 & ~ & ~ &   \\ \hline
        \multirow{9}{*}{L8} & \multirow{3}{*}{ZH} & EN & EN & EN & 44.72 & 22.16 & 37.16  \\ 
        ~ & ~ & EN & EN-ZH & EN-ZH & 42.86 & 17.89 & 33.51  \\ 
        ~ & ~ & EN-ZH & EN-ZH & EN-ZH & 46.59 & 22.69 & \textbf{38.48}  \\ \cline{2-8}
        ~ & \multirow{3}{*}{KO} & EN & EN & EN & 37.64 & 23.18 & 33.46  \\ 
        ~ & ~ & EN & EN-KO & EN-KO & 40.24 & 19.12 & 32.96  \\ 
        ~ & ~ & EN-KO & EN-KO & EN-KO & 42.48 & 23.13 & \textbf{36.39}  \\ \cline{2-8}
        ~ & \multirow{3}{*}{JA} & EN & EN & EN & 37.26 & 16.89 & 30.02  \\ 
        ~ & ~ & EN & EN-JA & EN-JA & 40.89 & 13.78 & 29.34  \\ 
        ~ & ~ & EN-JA & EN-JA & EN-JA & 39.33 & 17.04 & \textbf{31.18}  \\ \hline
        \multirow{6}{*}{HR} & \multirow{3}{*}{ZH} & EN & EN & EN & 52.54 & 30.15 & 45.74  \\ 
        ~ & ~ & EN & EN-ZH & EN-ZH & 56.77 & 27.03 & 46.53  \\ 
        ~ & ~ & EN-ZH & EN-ZH & EN-ZH & 54.89 & 30.35 & \textbf{47.25}  \\ \cline{2-8}
        ~ & \multirow{3}{*}{KO} & EN & EN & EN & 28.31 & 15.37 & 24.23  \\ 
        ~ & ~ & EN & EN-KO & EN-KO & 31.10 & 12.97 & 24.31  \\ 
        ~ & ~ & EN-KO & EN-KO & EN-KO & 29.77 & 15.57 & \textbf{25.18}  \\ \hline
        \multicolumn{2}{c}{\multirow{7}{*}{BEA-19}} ~ & EN & EN & EN & 58.32 & 35.20 & 51.55  \\ 
        ~ & ~ & EN & EN-ZH & EN-ZH & 57.13 & 30.92 & 48.85  \\ 
        ~ & ~ & EN-ZH & EN-ZH & EN-ZH & 59.02 & 36.21 & \textbf{52.42}  \\ 
        ~ & ~ & EN & EN-KO & EN-KO & 56.24 & 31.01 & 48.37 \\ 
        ~ & ~ & EN-KO & EN-KO & EN-KO & 57.46 & 36.94 & 51.72  \\ 
        ~ & ~ & EN & EN-JA & EN-JA & 57.87 & 30.12 & 48.87  \\ 
        ~ & ~ & EN-JA & EN-JA & EN-JA & 58.41 & 36.75 & 52.25 \\ \hline

    \end{tabular}
    \caption{Table showing the performance of the models trained with and without data augmentation on Stage 1 compared to the baseline model on the Lang-8 (L8), Human Re-annotated (HR) and BEA-19 Dev datasets. Note that all the experiments were conducted on a single seed using their respective CSW models. }
    \label{tab:stage1}
\end{table*}

We first observe that, surprisingly, in the EN-ZH test sets (both original and re-annotated), almost all span selection methods improve performance over the baseline in all combinations of training stages. This might suggest that the span selection method is not important and it is enough simply to expose a model to CSW. That said, the \verb|noun-token| and \verb|rand-phrase| methods consistently improve upon the baseline in almost \textit{all} settings. This might instead suggest that synthetic CSW data is more effective as GEC training data when it is linguistically informed. 

To explore this hypothesis more carefully, we can compare the results for \verb|cont-token| and \verb|ratio-phrase|, which both select spans of similar lengths (based on the average number of CSW tokens in the dataset) -- the former selects spans randomly while the latter selects only complete syntactic phrases. Although neither method consistently improves upon the baseline, the syntactically constrained \verb|ratio-phrase| method typically scores higher than the random \verb|cont-token| method. This leads us to conclude that linguistic insight is an important factor in synthetic CSW generation. 

Additionally, the \verb|rand-phrase| method also performed more consistently compared to the \verb|ratio-phrase| method which had a higher statistical similarity to the test set despite both methods having similar linguistic plausibilities. This may be down to the lack of CSW ratio variation across the sentences. However, more work has to be done in this area to identify the effect of CSW ratio on performance.

Nevertheless, the high variation in performance between the languages suggests that different languages respond to the span selection methods in different ways. This may be due to the varying placement of CSW points for different language combinations (cf. \S\ref{related:cs:gen}). On the other hand, the high variation between the different seeds for some of the methods (notably \verb|ratio-token| and \verb|cont-token|) is likely due to the difference in the tokens selected for translation. This highlights the sensitivity of such models towards the linguistic accuracy of CSW text used during training. 

Ultimately, we select the \verb|noun-token| method as the best method to use in all future experiments given that it improved upon the baseline in all but one test set (EN-KO). We do note, however, that \verb|rand-phrase| was also a strong contender given that it only struggled with EN-JA.

\subsection{Stage 1 Training Data}
\label{results:stage1}

Having identified \verb|noun-token| as the most promising span selection method for synthetic CSW generation, we next investigated whether it should also be applied to the Stage 1 training data. While our previous experiment did not reveal much difference in terms of whether we applied our method to just Stage 2, just Stage 3, or both, we nevertheless applied our method to both Stage 2 and Stage 3 in this experiment.

\autoref{tab:stage1} hence shows that applying \verb|noun-token| to all three training stages yields the best results for all our CSW test sets. This is expected since this setup exposes the model to the largest amount of CSW text during fine-tuning. However, it is surprising to see that the model trained using an EN Stage 1 dataset sometimes performs worse than the baseline model (notably the EN-ZH test set), despite being trained on CSW text in Stage 2 and 3. This might reflect the noisy nature of the original Lang-8 test sets, as the same pattern is not observed in the human re-annotated Lang-8 test sets.

To further investigate whether our CSW extensions affected the performance on monolingual GEC datasets, we also evaluated all these models on the monolingual English BEA-19 Dev dataset. \autoref{tab:stage1} thus also shows that the models trained with 3-stage CSW augmentation on all three languages did not negatively impact monolingual performance, and even brought about small improvements. The models trained on the monolingual EN Stage 1 dataset, however, all performed worse than the baseline, which might suggest the models learnt to focus too much on CSW in Stage 2 and Stage 3.

\begin{table}[!t]
    \centering
    \begin{tabular}{llcccc}
    \hline
        \multicolumn{2}{c}{Test} & \multicolumn{4}{c}{Training Dataset} \\
        \multicolumn{2}{c}{Dataset} & EN-ZH & EN-KO & EN-JA & EN \\ \hline
        \multirow{3}{*}{L8} & ZH & \textbf{38.48} & 37.87 & 36.81 & 37.16  \\ 
        ~ & KO & 36.08 & \textbf{36.39} & 36.25 & 33.46  \\ 
        ~ & JA & 31.22 & \textbf{31.28} & 31.18 & 30.02  \\ \hline
        \multirow{2}{*}{HR} & ZH & \textbf{47.25} & 46.68 & 46.61 & 45.74  \\ 
        ~ & KO & 22.98 & \textbf{25.18} & 23.19 & 24.23  \\ \hline
        \multicolumn{2}{l}{BEA-19} & \textbf{52.42} & 51.72 & 52.25 & 51.55 \\ \hline
    \end{tabular}
    \caption{$F_{0.5}$ score of the CSW models tested on the Lang-8 (L8), Human Re-annotated (HR) and BEA-19 Dev datasets. Only a single seed is used to produce the results shown.}
    \label{tab:inter_lang}
\end{table}

\subsection{Cross-Lingual Transferability}
\label{results:language_inter}

To investigate the transferability of models beyond the CSW languages they were trained on, we evaluate each CSW model (trained with CSW data in all three stages) on all the test datasets. \autoref{tab:inter_lang} thus shows that the EN-ZH and EN-KO models perform best on the EN-ZH and EN-KO test sets respectively, as expected, but that the EN-KO model slightly outperforms the EN-JA model on the EN-JA data, albeit by a very small margin ($\sim$0.10 $F_{0.5}$).

The CSW models also outperform the monolingual model in almost all testing scenarios (except the human re-annotated EN-KO test set), which suggests that it is generally better to have a CSW GEC model than a monolingual model, even if the CSW GEC model is trained on a different language pair to the intended use-case. This intuitively makes sense, as a model likely benefits from having an explicit concept of `other' language even if that language is different from what it was trained on. Nevertheless, it is expected that the best results will come from training data that most closely resembles the target CSW language pair. 

It is finally worth mentioning that this effect may also be influenced by the similarity of the languages in question. For example, Japanese and Korean share a similar word order, while Japanese and Chinese share a subset of characters. We thus hypothesise that comparable linguistic features may have an effect on the success of CSW GEC. 

\subsection{Linguistic Plausibility}

To explore this idea further, we also consider how plausible our synthetic CSW datasets are. In particular, given that both Japanese and Korean are head-final languages, while English is predominantly head-initial, we might expect some constraints given their disharmonious word-orders. In these cases, methods such as \verb|ratio-token| or \verb|cont-token| might not yield output that would be considered realistic. Example (\ref{ex:err}) demonstrates some typical cases. 

\pex \label{ex:err}
Source: `I think that public transport will always exist in the future .' 
\a \label{ex:enjaerror} 
\begingl[everygla=,everyglb=,aboveglbskip=-.2ex, aboveglftskip=.1ex]
\gla I think that public transport \Japanese{意思} always \Japanese{存在} in the future . //
\glft {} 
\rightcomment{\textbf{[EN-JA ratio-token]}} //
\endgl

\a \label{ex:enkoerror}
\begingl[everygla=,everyglb=,aboveglbskip=-.2ex, aboveglftskip=.1ex]
\gla I think that public transport will \Korean{항상 존재한다} in the future .//
\glft {}
\rightcomment{\textbf{[EN-KO cont-token]}}//
\endgl
\xe

\noindent In both cases, the verb `exist' (\Japanese{存在} in Japanese, \ref{ex:enjaerror} and \Korean{항상 존재한다} in Korean, \ref{ex:enkoerror}) would be expected to come at the end of the sentence in their respective languages. These utterances would thus not be considered plausible according to \citeauthor{poplack_syntactic_1978}'s (\citeyear{poplack_syntactic_1978}) Equivalence Constraint (\textsection \ref{related:cs:gen}), for example. Similarly, since spans are translated out of context, there is also a chance that the wrong translation is generated. This is what happens in (\ref{ex:enjaerror}), where the English modal verb `will' has been confused with the noun meaning `intention' and translated into the Japanese noun \Japanese{意思} accordingly, representing a generated switch that is both syntactically and semantically nonsensical. This might explain why some span selection methods performed worse than others, especially random selections. Without human judgements and further experiments, it remains unclear how close the synthetic constructions are to reality, as well as the extent to which linguistic plausibility impacts systems' performance.

\section{Conclusion}
In this paper, we conducted the first study into developing GEC systems for CSW text. We specifically investigated the performance of various pre-trained models on CSW text and compared different methods of generating synthetic CSW GEC data to improve the performance of the sequence-tagger-based GEC model GECToR. This was achieved by: 
\begin{enumerate}[i.]
    \item automatically translating different spans of text within existing GEC corpora; and
    \item assessing the performance of the models on a subset of the multilingual Lang-8 dataset which we reannotated and release with this paper.
\end{enumerate}

\noindent Our findings suggest that data augmentation is most effective in the context of multilingual (rather than monolingual) pre-trained models (e.g.\ XLM-RoBERTa). Moreover by experimenting with different methods of generating synthetic CSW GEC datasets, it was found that replacing a random noun token in each sentence yielded the best improvement compared to other methods and the baseline. This finding is consistent with observations in linguistics, given that the most linguistically motivated method yielded the best improvement, and highlights the potential contribution of linguistic insights when building tools to process CSW text. 

We also found that applying our data augmentation method to all 3 stages of the GECToR training process returned the best results over the baseline. This may be unsurprising, but is consistent with the original GECToR finding that all training stages have a significant impact on the final model performance. Finally, we also discovered that models trained on one CSW language generalise relatively well to other CSW languages and even improve performance on a benchmark monolingual dataset. This suggests that multilingual transfer can be used to improve an out-of-domain CSW GEC system when in-domain CSW GEC data is not available. Ultimately, we have shown that data augmentation improves the model's performance on CSW text both on the Lang-8 dataset and a human re-annotated subset. These results lay important foundations for future work.

\section{Future work}
We conclude by identifying several possible directions for future work. 

First, inspired by \citet{chang_code-switching_2019}, we would suggest using a Generative Adversarial Network (GAN) \citep{goodfellow_generative_2014} to predict and generate CSW points for the different languages. Although we found that our \verb|noun-token| method is most effective at improving the performance of GEC models on CSW text, this method does not fully consider all different linguistic constraints and the varying switching points in different languages. It is thus worth considering other methods of CSW generation which may be more effective than the \verb|noun-token| method used in this project.

Second, it is worth exploring the development of multilingual CSW models trained on a combination of CSW texts from different languages. In this work, we developed a separate model for each CSW dataset, but this is impractical when you consider the number of possible CSW combinations. It may thus be effective to combine many different CSW combinations in a single dataset.

Third, it would also be useful to investigate whether our data augmentation methods are more, or less, effective in the context of machine translation-based GEC models. With the recent rise of generative language models, the performance of machine translation on these models has greatly improved \citep{hendy_how_2023}. 

Furthermore, there is a huge potential to extend this method to GEC on code-switching in other language pairs. Although sufficient annotated data for each pairing remains an issue \citep{dogruoz-etal-2021-survey,dogruoz-sitaram-2022-language,dogruoz-etal-2023-representativeness}, our data generation pipeline is language-agnostic and could thus be applied more widely to aid progress in this direction. In fact, current computational work on code-switching remains generally biased towards major languages such as English/Spanish or English/Hindi (e.g. \citealp{srivastava_overview_2022,chen_calcs_2022,Nguyen-2021}, i.a. -- see also \citealp{dogruoz-etal-2021-survey,dogruoz-sitaram-2022-language,dogruoz-etal-2023-representativeness} for a comprehensive overview), and so there is a lot to explore with more diverse datasets. 

Finally, making use of the enormous amount of work in Linguistics on code-switching should also be a focus for future work (cf. \citealp{dogruoz-etal-2021-survey,dogruoz-sitaram-2022-language}). As we discovered in this study, a large portion of CSW sentences contain single-token-long CSW components, most of which were nouns. These findings are in line with what has long been known in Linguistics, and highlight the potential of incorporating such insights into building the next generation of tools to process CSW input. This also motivates further exploration of different features which may aid the production of synthetic CSW datasets. 

\section*{Limitations}

Our data generation method depends on Google Translate, which is a closed-source service provided by Google. It is unclear how frequently this service is updated; this dependency adds variability when replicating our results.  

\section*{Ethical Considerations}

The work reported in this paper was undertaken in an ethical manner. Specific points to highlight:

\begin{itemize}
    \item Our human annotators are colleagues who were approached on a volunteer basis. They were under no obligation to assist us, but did so as voluntary research contributions.
    \item To save energy, we did not train all possible combinations of models on GPUs, but defined a strategy to only explore the most promising (cf.\S\ref{proposed}). 
\end{itemize}

\section*{Acknowledgements}
We thank Fanghua Zheng and Oliver Mayeux for their generous help in re-annotating the English-Chinese and English-Korean datasets, respectively.
This work was supported by Cambridge University Press \& Assessment.

\section*{Bibliographical References}
\label{reference}
\bibliography{custom.bib}
\bibliographystyle{lrec-coling2024-natbib}

\end{document}